\pdfoutput=1

\documentclass[11pt]{article}

\usepackage[preprint]{acl}
\usepackage{amsmath}
\usepackage{times}
\usepackage{latexsym}
\usepackage{makecell}
\usepackage{multirow}
\usepackage{colortbl}

\usepackage{caption}
\usepackage{subcaption}
\usepackage[T1]{fontenc}

\usepackage[utf8]{inputenc}

\usepackage{microtype}

\usepackage{inconsolata}

\usepackage{graphicx}
\title{FCoT-VL:Advancing Text-oriented Large Vision-Language Models with Efficient Visual Token Compression}

\author{
  \textbf{Jianjian Li \textsuperscript{1}},
  \textbf{Junquan Fan\textsuperscript{1}},
  \textbf{Feng Tang\textsuperscript{2}}\thanks{corresponding author},
  \textbf{Gang Huang\textsuperscript{2}}\footnotemark[1],
  \textbf{Shitao Zhu\textsuperscript{2}},
  \textbf{Songlin Liu\textsuperscript{2}} \\
   \textbf{Nian Xie\textsuperscript{2}},
  \textbf{Wulong Liu\textsuperscript{2}},
  \textbf{Yong Liao\textsuperscript{1}\footnotemark[1]}
\\
  \textsuperscript{1}University of Science and Technology of China.\\
  \textsuperscript{2}Huawei Noah’s Ark Lab.\\
\\
}

\begin{document}
\maketitle
\begin{abstract}
The rapid success of Vision Large Language Models (VLLMs) often depends on the high-resolution images with abundant visual tokens, which hinders training and deployment efficiency. Current training-free visual token compression methods exhibit serious performance degradation in tasks involving high-resolution, text-oriented image understanding and reasoning.
In this paper, we propose an efficient visual token compression framework for text-oriented VLLMs in high-resolution scenarios. In particular, we employ a light-weight self-distillation pre-training stage to compress the visual tokens, requiring a limited numbers of image-text pairs and minimal learnable parameters. Afterwards, to mitigate potential performance degradation of token-compressed models, we construct a high-quality post-train stage.
To validate the effectiveness of our method, we apply it to an advanced VLLMs, InternVL2. Experimental results show that our approach significantly reduces computational overhead while  outperforming the baselines across a range of text-oriented benchmarks. We will release the models and code soon.

\end{abstract}

\section{Introduction}

The success of Large Language Models (LLMs) \cite{achiam2023gpt,yang2024qwen2,zhu2023minigpt,dubey2024llama3,bi2024deepseek,cai2024internlm2} has inspired efforts to extend their capabilities to other modalities, particularly vision. In vision-language tasks, VLLMs process visual features extracted from vision transformers (ViTs) \cite{radford2021clip}) and integrate them to LLMs. The performance of these models is often positively correlated with visual resolution.

Improving visual resolution in ViTs involves fixed high-resolution settings (e.g., CogVLM2 \cite{hong2024cogvlm2}, GLM4V9B \cite{glm2024chatglm}), slicing patch schemes (e.g., LLaVA 1.6 \cite{liu2024llavanext}, InternVL series \cite{chen2024far}), or simple dynamic resolution (Qwen2-VL \cite{wang2024qwen2vl}). These strategies enhance fine-grained visual understanding in models. However, higher resolutions drastically increase token count, imposing significant computational burdens. For example, Qwen2-VL processes 11,427 visual tokens for an image with a resolution of $8204 \times 1092$ pixels. This results in considerable computational overhead during both the training and inference phases, making high-resolution processing resource-intensive and challenging to scale-up. 


\begin{figure}[t]
\includegraphics[width=0.48\linewidth]{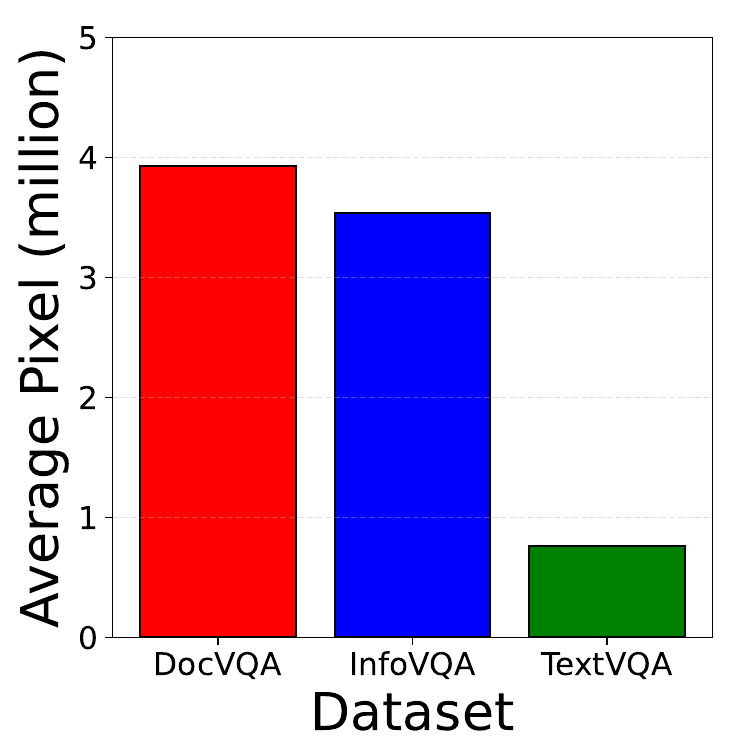}
    \includegraphics[width=0.48\linewidth]{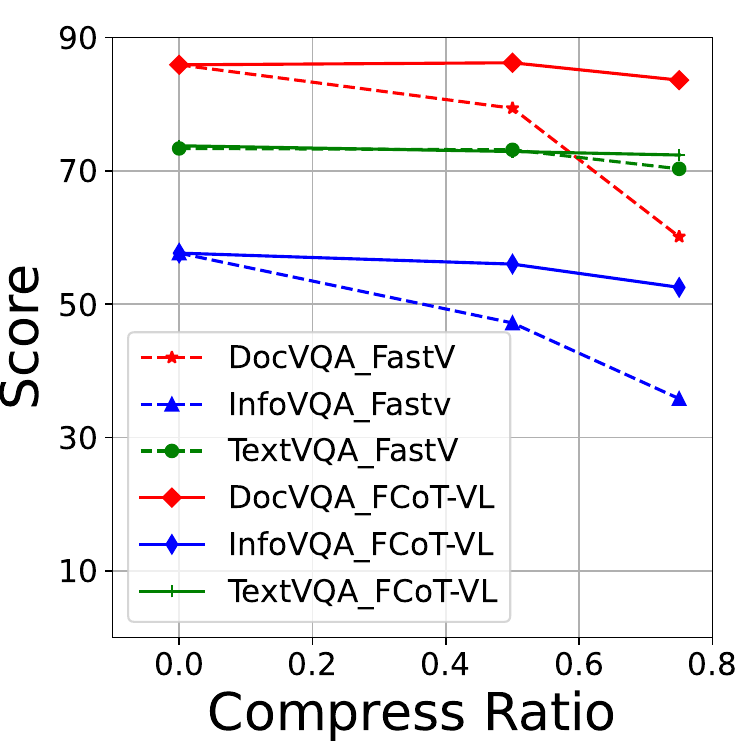}
    \caption {Comparison of scores between FastV and FCoT-VL on different types of benchmarks. FastV gets a significant decline in tasks that require high resolution like DocVQA and InfoVQA. In contrast, our method shows a minor performance degradation.}
    \label{fig:fastv}
\end{figure}

To resolve above issues, reducing visual tokens in well-trained VLLMs has been studied in works like LLaVA-PruMerge \cite{shang2024llava}, SparseVLM \cite{zhang2024sparsevlm} and VisionZip \cite{yang2024visionzip}. For instance, VisionZip\cite{yang2024visionzip} selects informative tokens using attention scores to reduce the total number of tokens. However, training-free token pruning methods, like FastV \cite{chen2025image} in Figure~\ref{fig:fastv}, shows sub-optimal performance in text-oriented tasks that demand high-fidelity token representations. To this end,
training from scratch with reduced visual tokens is another alternative. For example, TextHawk2 \cite{yu2024texthawk2} uses 100M pre-training image-text pairs to train cascaded decoder layers, progressively downsampling visual tokens by $4\times$ ratio. TextHawk2 requires significant data and resources, posing challenges in low-resource settings. 
This raises a challenge: \textbf{Can we compress the visual tokens effectively under constraints of limited training datas and GPUs resources?}

For this challenge, we \textbf{F}ocus on \textbf{Co}mpression of visual tokens in high-resolution \textbf{T}ext-oriented Large \textbf{V}ision-\textbf{L}anguage Models (FCoT-VL) while retaining fine-grained image detailed perception. To be special, we 
propose a self-distilling framework as shown in Figure~\ref{fig:Overvieww}, comprising a teacher model with abundant visual tokens and a student model with compressed token representations.
To build upon established capabilities, we adopt the InternVL2 to initialize the teacher model and student model. During the self-distillation process, only a lightweight token compression module and projector in the student model are learnable with a small-scale set of image-text pairs(i.e., 2M).
This approach brings two advantages: 1). The student inherits the parameter from the teacher, avoiding large-scale training and preserve advanced capabilities of the teacher model. (2) We exclusively finetunes the token compression module, which can achieve promising performance even with limited training data.

In practice, we find distilled student model has performance drops(about 5$\%$) inevitably.
To enhance the performance of the student model relative to the teacher model, we introduce a post-training stage using high-quality instruction datasets, involving documents, mathematics, science, charts, and GUI images. Besides, we propose a multi-stage model fusion technique that iteratively merges models to improve adaptability across various tasks. The post-training improves the model’s ability to handle complex tasks, such as document parsing and reasoning-based QAs. 

Our contributions can be concluded as follows:
\begin{itemize}
    \item [(1).]
    We propose a self-distilling paradigm towards visual token cimpressing for high-resolution text-oriented VLLMs, enabling robust re-alignment while minimizing both data and computational demands.
    \item [(2).]
    We explore post-training strategies including synthesis of high-quality supervised fin-tuning data and training-free model merging schemes, facilitating the capabilities of compressed VLLMs.
    \item [(3).]
    We develop the proposed FCoT-VL in the InternVL2 series, achieving compression ratios of 2 and 4, respectively. Extensive empirical evaluations across multiple text-oriented benchmarks reveal that our proposed models achieve comparable or superior performance to existing token-rich VLLMs, while offering higher training and deployment efficiency.
\end{itemize}

\section{Related Works}
\subsection{Vision Large Language Models}


In recent years, open-source  VLLMs have made significant advancements, driven by contributions from both academia and industry. 
Earlier models, such as BLIP-2 \cite{li2023blip}, MiniGPT\cite{zhu2023minigpt} and LLaVA\cite{liu2024visual,liu2024llava}, have proven to be effective for vision-language tasks via bridging off-the-shelf ViTs and LLMs. 
However, early VLLMs struggle with processing images containing
fine-grained details, especially for OCR-like tasks such as charts\cite{masry2022chartqa}, documents\cite{mathew2021docvqa},
and infographics\cite{mathew2022infographicvqa}. To this end, InternVL series propose
an adaptive cropping method to convert vanilla images as several fixed image patches. For example, InternLM-XComposer2-4KHD\cite{dong2024internlm} increases 336 pixels of CLIP to 4K resolution and gets strong document understanding ability. InternVL2
obtains promising results on text-oriented benchmarks via scaling up image resolution and ViT model parameters. Moreover, QwenVL2 \cite{wang2024qwen2vl} proposes a native dynamic processing of images at varying resolutions. This image processing setting generates more visual tokens and suppress adaptive cropping VLLMs. However,
high-resolution processing pipelines bring substantial computational overhead in both training and inference stages, hindering real-world deployment.


Beyond high-resolution tricks, many works reveal that high-quality datas are more important for advancing document understanding. Recent studies\cite{hu2024minicpm,li2024baichuan,li2025eagle} highlight the critical role of data quality in VLLMs. For instance, InternVL-2.5\cite{chen2024expanding} enhanced performance of previous version through collecting more diverse dataset and data processing pipelines.

In this paper, we also explore how to obtain high-quality post-training datas to match frontier open-source VLLMs. Specifically, Our FCoT-VL outperforms the base model InternVL2
on many benchmarks like ChartQA\cite{masry2022chartqa} and MathVista\cite{lu2024mathvista}, despite reducing visual tokens by 50$\%$.

\begin{figure*}[t]
  \centering
  \includegraphics[width=\textwidth]{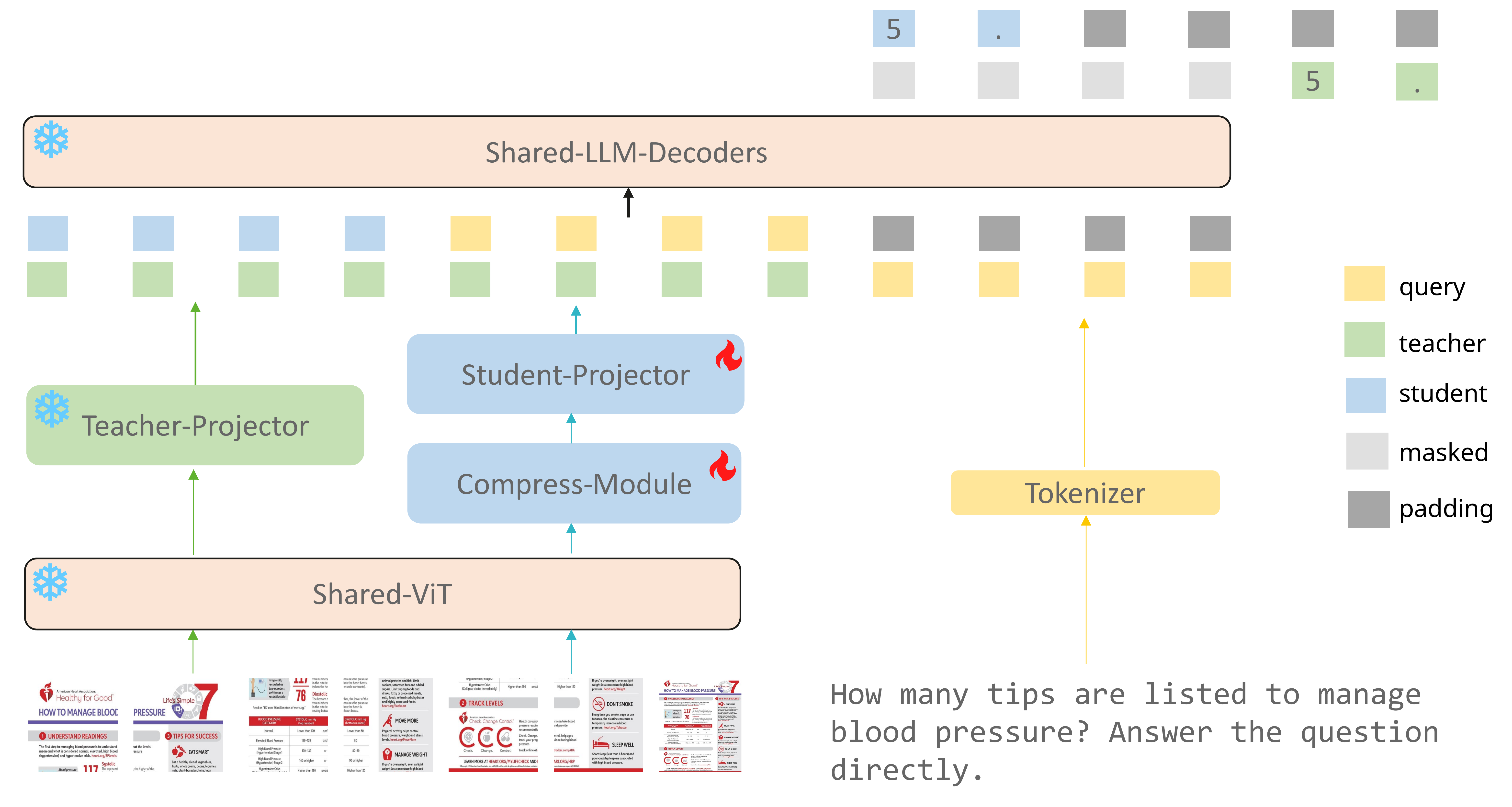}
  \caption{Overall Structure of FCoT-VL. FCoT-VL is a self-distillation architecture in which only the Student-Projector and Compress-Module are learned, while all the other modules remain frozen. The student and teacher models share the same ViT encoder and the LLM decoder.}
  \label{fig:Overvieww}
\end{figure*}
\subsection{Visual Compression Schemes}
Visual compression, a key focus in high-resolution VLLMs, aims to efficiently reduce the use of vision tokens, minimizing computational and memory overheads. The inherent redundancy of visual data, compared to dense textual data, underscores the importance of compression.

Solutions to visual compression can be broadly categorized into two main approaches: training-free and training-based ones. Training-free methods dynamically select more important vision tokens via various strategies during decoding stage. For instance, SparseVLM\cite{zhang2024sparsevlm} and VisionZip \cite{yang2024visionzip} prioritize tokens based on attention scores. ToMe\cite{bolya2022token} and LLaVA-PruMerge\cite{shang2024llava} cluster tokens using cosine similarity. However, the training-free paradigms suffer from significant performance drops in text-orientated benchmarks. 
In contrast, training-based methods focus on optimizing the visual adaptor by incorporating external modules for token reducing. For instance, LLaMA-VID\cite{li2024llama} enhances visual information extraction through Q-Former\cite{li2023blip} with context tokens. Similarly, models like C-Abstractor\cite{cha2024honeybee} and LDP\cite{chu2024mobilevlm} serve as promising alternatives for visual token compressing.

Training from scratch necessitates extensive alignment datasets and substantial computational resources, often consuming thousands of GPU days. In this work, we present an efficient training token-compressing framework that achieves comparable performance while significantly reducing both data and computational requirements.

\section{Method}
We propose FCoT-VL, a framework for compressing visual tokens in VLLMs. It has the following objectives: (1). The efficient realignment training stage. we propose a self-distillation framework to transfer visual token knowledge from rich-token VLLM to compressed-token VLLM. We only learn lightweight parameters with limited datas to acquire visual token compression ability while maintaining training and inference efficiency. (2). To boost text-oriented VLLM after visual token cutoff, we focus on advanced post-training and data augmentation techniques, enabling the student model to catch up with InternVL2.

\subsection{Architecture of FCoT-VL}
As shown in Figure~\ref{fig:Overvieww}, we present the architecture design of our FCoT-VL, which comprises a vanilla VLLM as the teacher model(i.e., InternVL2) and a VLLM with compressed visual tokens as the student model in the distillation process.
\subsubsection{Re-alignment}
\paragraph{Definition} As illustrated in Figure~\ref{fig:Overvieww}, the basic architecture of the re-alignment consists of five primary components: a shared visual encoder $ViT_{\phi}$, a shared large language model $LLM_{\theta}$, a teacher visual adaptor $A_t$ and a student visual adaptor $A_s$, and a visual token compression module $V_c$. Given a visual instruction input ($x_t$, $x_v$, $y$), then the responses are computed as follows:
\begin{equation}
    \left\{
    \begin{aligned}
        \hat{y_t} & = LLM_{\theta}[A_{t}(x_v);x_t] \\
        \hat{y_s} & = LLM_{\theta}[A_{s}(V_{c}((x_v);x_t]
    \end{aligned}
    \right.
\end{equation}
where $[;]$ means the concatenation operation, $x_v$ is the input image and $x_t$ is the text instruction embeddings. The $\hat{y_t}$ and $\hat{y_s}$ denote the probabilities of  responses for the teacher model $t$-VLLM and 
student model $s$-VLLM, respectively.
\paragraph{Initialization} We initialize our student model inherited from the teacher model parameters. During re-alignment stage, we freeze all the parameters of the teacher models. The $LLM_\theta$ and $ViT_{\phi}$ in $s$-VLLM maintain frozen since their pre-trained parameters have already captured rich visual and language knowledge. Only the student adaptor $A_s$ and the visual token compression module $V_c$ are learnable to  bridge different modalities and compress visual tokens of the LLM part.

\paragraph{Self-distillation} We compare different visual token adjustment methods like Qformer \cite{li2023blip}, pooling and convolution as $V_c$ as in Table~\ref{tab:different_adapter}. We find that
employing a simple convolutional layer could reduce visual tokens ($\times$4 and $\times$2 ratio) effectively.


We aim to re-align the visual tokens with text tokens in the $s$-VLLM using OCR-like tasks, which converts texts in the images into an editable text format. Different from previous training distilling methods, which focuses on QA tasks, we argue that OCR-like tasks  require models to perceive dense information of the whole image and benefit efficient re-alignment for FCoT-VL. Accordingly, our OCR data sources are sampled from $t$-VLLM(i.e., InternVL2 series) with a small amount 2M image-text pairs, covering text recognition, layout parsing from web, natural, document, table, chart and handwritten samples. 

To maintain and leverage the performance of the teacher model,
the training objective is to minimize the Kullback-Leibler (KL) divergence between the output logits of $t$-VLLM and $s$-VLLM. The objective function is:
\begin{equation}
    \mathcal{L}_{\text{KL}}(\hat{y_t} \parallel \hat{y_s}) = \sum_{i}^{N} \hat{y_t}(i) \log \left( \frac{\hat{y_t}(i)}{\hat{y_s}(i)} \right)
\end{equation}
Where $\hat{y_t}$ and $\hat{y_s}$ are the logits of the teacher model and student model, receptively. $N$ is the total token length. The output of the teacher model plays the role of soft labels to guide visual token compression. Additionally, we find that introducing ground-truth answers as hard labels contributes to stable training. The the Cross-Entropy loss is:
\begin{equation}
    \mathcal{L}_{\text{CE}} = -\sum_{i}^{N}\log\hat{y_s}(i))
\end{equation}
Then the optimization goal is to minimize the $\mathcal{L}=\mathcal{L}_{\text{KL}} + \mathcal{L}_{\text{CE}}$.

\subsubsection{Post-Train}\label{sec:pt}
In this section, we describe supervised fine-tuning(SFT) aimed at improving the student model's performance in text-oriented tasks. We accept many open-source datasets reported in previous VLLMs \cite{chen2024far}, covering a variety of downstream tasks. However, we find that many of these public datasets are not formatted in an instruction style. To overcome this, we leverage distillation from teacher models to acquire the conversation style. Subsequently, we prompt the InternLM2.5-7B \cite{cai2024internlm2} to rewrite the instruction datas with the tone of the teacher model. Moreover, we observe this rewriting method facilitates fast and stable training, which may be attributed to the strong alignment with the teacher model.

\paragraph{Chain-of-Thought pipeline.} \label{abnormal}For reasoning tasks like math, chart reasoning and calculation problems, we leverage Rejection Sampling (RS) to expand the SFT dataset using larger and stronger multimodal language models. Specifically, for the question $q$, we employ RS to generate a new response with COT, obtaining the reasoning steps $R_{cot}$ and final answer $R_{ans}$, respectively. We use rule-based verifications that verify the correctness of the concluded answer $R_{ans}$ for the given problem $q$ based on the ground truths. We find that the mixture of RS-augmented and vanilla data significantly enhances reasoning capabilities. For example, our FCoT-VL-2B, with half visual tokens retained, achieves a score of 58.96 on MathVista \cite{lu2024mathvista}, outperforming many 7B-scale VLLMs.

\paragraph{Data sampling pipeline.} Considering that our tasks cover diverse image understanding and reasoning tasks with varying difficulty levels in a single SFT stage, we develop a novel sampling strategy, termed post-training sampling, to address these potential issues. Specifically, we perform coarse training using a small subset of the entire dataset at first, and then analyze the training loss distributions across different tasks. For datasets exhibiting much lower loss values, indicating easier learning, we down-sample them in the subsequent formal training. Conversely, we identify tasks (excluding generation tasks) with higher loss values and increase their sampling probabilities, addressing the model's weaknesses, especially in reasoning tasks.
 


                 
                 



\paragraph{Model Merging.}
Since our SFT training covers many tasks,  we aim to merge the base model with weighted differences from each checkpoint during training. These checkpoints reflect different stages of fine-tuning, with each stage capturing important task-specific adaptations. During training, multiple intermediate checkpoints are saved, and they are merged using the following formula:
\begin{equation}
M_{\text{mge}} = \theta_{\text{base}} + \sum_{i=1}^{n} \alpha_i (\theta_{\text{cpt}_i} - \theta_{\text{base}})
\end{equation}
Where $M_{\text{mge}}$ is the merged model, \(\theta_{\text{base}}\) is typically used as the final model, and \(\alpha_i\) is the weight for the difference between the checkpoints $\theta_{\text{cpt}_i}$ and the base model. $n$ is set as 5.
The goal is to determine the optimal fusion weights, formulated as:
\begin{equation}
\arg\max_{\alpha_1, \dots, \alpha_n} f( \theta_{\text{base}} + \sum_{i=1}^{n} \alpha_i (\theta_{\text{cpt}_i} - \theta_{\text{base}}))
\end{equation}
Rather than relying on costly heuristic algorithms, we use Shapley values\cite{sundararajan2020many}, to fairly serve the merge weight \(\alpha_i\) to each checkpoint \(M_i\) based on its contribution to the final model performance. The weighted combination of checkpoints thus optimizes the final model's performance based on their individual contributions.

\paragraph{Computation Complexity.} 
In this section, we analyze the computation complexity of FCoT-VL in our post-training stage. The computational burden in the FCoT-VL is predominantly attributed to the attention operations within the LLM decoders. Assuming the LLM decoders has \( L\) layers, we only compute the complexity of one self-attention and feed-forward network, yielding: 
\begin{equation}
O(L \cdot (n^2 \cdot d + n \cdot d^2))
\end{equation}
Where $n$ is the length of input vectors and $d$ is the dimension of LLM's input tokens. When the compress ratio is $r$, the computation complexity could be reduced as:
\begin{equation}\label{eq_7}
O(L \cdot (\frac{n^2 \cdot d}{r^2}  + \frac{n \cdot d^2}{r}))
\end{equation}

Since the computation cost of LLM decoders is dominant in the our FCoT-VL, the overall computation complexity will be reduced much, facilitating training and inference effectiveness. More quantitative experiments are discussed in Section~\ref{timecost}.

\begin{table*}
    \setlength{\tabcolsep}{1.5pt}
    \renewcommand{\arraystretch}{1.5}
    \centering
    \resizebox{\textwidth}{!}{
        \begin{tabular}{lc*{9}ccc|c}
            \hline
             \textbf{Base Model} & \textbf{Method} & \textbf{\makecell{Compress \\Ratio}}  & \textbf{DocVQA} & \textbf{ChartQA} & \textbf{TextVQA} & \textbf{InfoVQA} & \textbf{OCRBench} & \textbf{\makecell{OCRBench \\ v2 En}} & \textbf{\makecell{OCRBench \\ v2 Ch}} & \textbf{AI2D} & \textbf{MathVista} & \textbf{ScienceQA} & \textbf{\makecell{Avgs \\($\%$)}} \\
             \hline
             \multirow{2}*{\makecell{LLaVA-1.5 \\ 7B}}  &original & 0\% & 28.10 & 17.8 & 58.2 & 25.8 & - & - & - & 55.5 & 25.6 & - &  100 \\
             \cline{2-14}
             ~  &\multirow{1}*{FastV}& 50\% & -  & 17.7 & 45.5 & - & - & - & - & - & - & - & 88.81\\
             
             \hline
             \multirow{3}*{\makecell{LLaVA-NeXT \\ 8B}} &original & 0\% & 78.22 & 69.28 & 65.41 & - & - & 31.5& 9.1 & - & - & - & 100 \\
             \cline{2-14}
             ~  &\multirow{2}*{FastV}& 50\% & 73.92  & 67.60 & 65.15 & - & - & - & - & - & - & - & 97.23\\
             
             ~  & ~ & 75\% & 66.67 & 62.80 &  63.08 & - & - & - & - & - & - & - & 90.77 \\
             \hline
             \multirow{5}*{\makecell{InternVL2 \\ -2B}} &original & 0\% & 85.90 & 76.24 & 73.36 & 57.66 & 78.4 & 35.7 & 34.5 & 74.09 & 46.30 & 94.25 & 100 \\
             \cline{2-14}
             ~  &\multirow{2}*{FastV}& 50\% & 79.39 & 69.72 & 73.15 & 47.18  & 73.3 &33.2& 26.1 & - & - & - & 89.65\\
             ~  & ~ & 75\% & 60.12  & 60.76  & 70.3 & 35.82  & 64.3  & 29.0 & 19.8 & - & - & - & 75.47 \\
             \cline{2-14}
             ~  &\multirow{2}*{Ours}& 50\% & \textbf{86.21}& \textbf{78.46} & 72.90  & 56.01 & \textbf{80.2}  & 35.6  & \textbf{34.8}  & \textbf{85.80}  & \textbf{58.96}  & 90.68 & \textbf{104.20}\\
             
             ~  & ~ & 75\% & 83.60 & 75.84  & 72.37 & 52.52  & \textbf{81.2} & 33.5 & 34.4 & \textbf{84.20} & \textbf{52.90} & 91.72  & \textbf{100.91} \\
             \cline{2-14}
            \hline
              \multirow{5}*{\makecell{InternVL2 \\ -8B}}&original & 0\% &91.6 & 83.3 & 77.4 & 74.8 & 79.4 & 39.6& 36.3 & 83.77 & 58.3 & 97.22 & 100\\
              \cline{2-14}
             ~  &\multirow{2}*{FastV}& 50\% & 85.25  & 79.5 & \textbf{77.61}& 60.9 & 76.1  & 36.8 & 26.4 & - & - & - & 93.21 \\
             
             ~ & ~ & 75\% & 67.52 & 73.52 & 74.65 & 46.06 & 68.1 &28.7 & 21.2 & - & - & - & 81.15\\
             \cline{2-14}
             ~  &\multirow{2}*{Ours}& 50\% & \textbf{91.88}  & \textbf{85.52} & \textbf{78.95} & 71.71 & \textbf{83.9}  & \textbf{42.1} & \textbf{40.1} & \textbf{93.80} & \textbf{63.3} & 95.14 & \textbf{103.43}\\
             
             ~  & ~ & 75\% & 89.91  & \textbf{84.16}  & \textbf{77.80}  & 67.11 & \textbf{82.0} & \textbf{41.8} & \textbf{36.7} & \textbf{93.48} & \textbf{62.00}  & 93.40  & \textbf{100.84} \\
             \hline
        \end{tabular}
    }
    \caption{Performance comparison across various text-oriented tasks under different compression ratios settings. This table summarizes the performance metrics of different models at different compression ratios. Our benchmarks include document, chart, natural, scientific and math images. Items that outperform the baseline are bolded in the table and the average performance across all tasks is also provided in the last column. }
    \label{tab:main_result}
\end{table*}

\section{Experiments}
To validate the effectiveness of FCoT-VL, we evaluate on nine text-oriented mutimodal benchmarks: 
DocVQA\cite{mathew2021docvqa}, ChartQA\cite{masry2022chartqa}, TextVQA\cite{singh2019towards}, AI2D\cite{kembhavi2016diagram}, InfoVQA\cite{mathew2022infographicvqa}, OCRBench\cite{liu2024ocrbench}, OCRBench\_v2\cite{fu2024ocrbench},  MathVista and ScienceQA\cite{lu2022scienceqa}. 

\subsection{Main Results}
We choose InternVL2-2B and InternVL2-8B as our baseline models, considering that their good adaption to high-resolution images and impressive performance.
As shown in Table \ref{tab:main_result}, we compress the visual tokens at ratio 50$\%$ and 75$\%$ of InternVL2-2B and InternVL2-8B, respectively. For the training-free FastV method,
we find significant performance drop on the different baseline VLLMs(i.e., LLaVA-1.5-7B\cite{liu2023improved}, LLava-NeXT\cite{liu2024llava} and InternVL2\cite{chen2024far}), particularly when the visual tokens drop to 1/4. 
For instance, at a compressing ratio of 50\%, the performance degradation is approximately 10\% on InternVL2-2B, but at 75\% compressing ratio, the performance drop exceeds 25\%.
This suggests that training-free
paradigm is insufficient in text-oriented tasks, specially for high-resolution and text-rich images.


As for our FCoT-VL, it achieves more than 100\% average performance over baselines at the 50\% and 75\% visual token compressing ratios, respectively. More surprisingly,
even under an extreme compressing ratio of 75\%, FCoT-VL-2B exhibits only a slight performance degradation of approximately 5\% across most benchmarks when compared to baselines, making it a compelling choice for low-resource deployment inference.

\begin{figure}[t]
\includegraphics[width=0.49\linewidth]{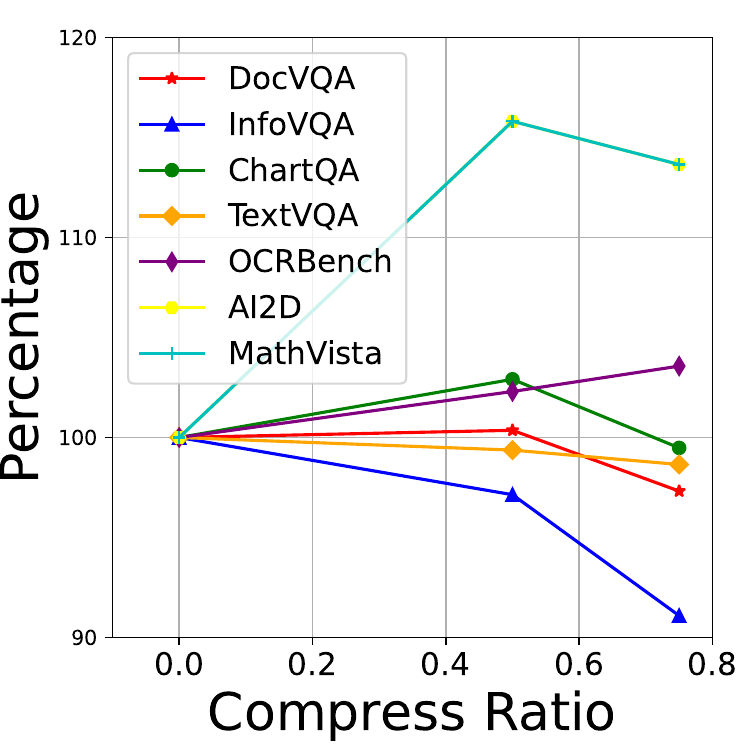}
\includegraphics[width=0.49\linewidth]{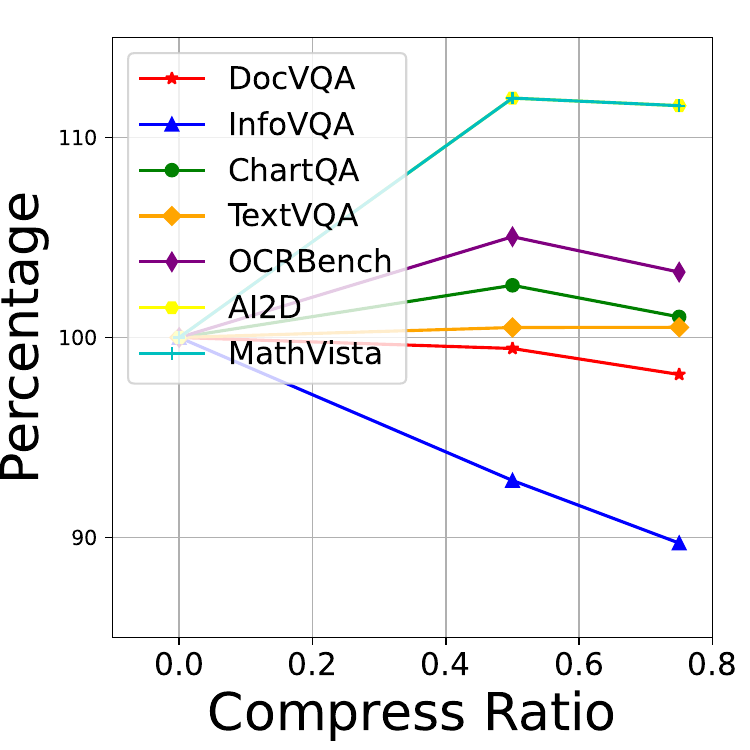}
  \caption {Performance percentage across multiple benchmarks under different compression ratios on the InternVL2-2B (left) and InternVL2-8B (right) models.}
  \label{fig:percentagefcotvl}
\end{figure}

Additionally, we visualize the percentage of performance variations across different tasks, as shown in Figure~\ref{fig:percentagefcotvl}. For DocVQA and InfoVQA, which heavily rely on high-resolution images, our FcoT-VL still inevitably incurs some performance degradation. 
This highlights the trade-off between token compressing and the performance fine-grained visual details in tasks demanding high-resolution inputs. In contrast,
FCoT-VL achieves performance improvements on tasks that demand advanced vision understanding and reasoning capabilities, such as OCRBench, OCRBench\_v2, AI2D, and MathVista. These obversations validate that high-quality data (as discussed in Section~\ref{abnormal}) plays a more critical role in enhancing performance than simply relying on resolution scaling laws.

\subsection{Ablation Study}\label{ablationstudy}

\begin{figure}[ht]
    \includegraphics[ width = \linewidth]{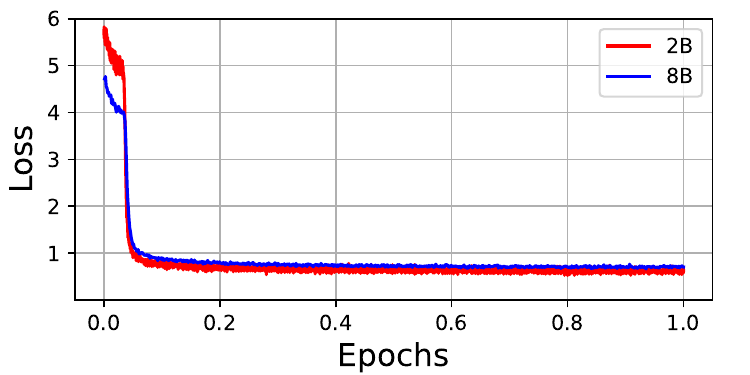}
    \caption {The loss graphs of re-alignment pre-training. The loss undergoes a rapid loss reduction and a long smooth convergence.}
    \label{fig:pretrain_loss}
\end{figure}

\paragraph{Re-alignment}
We implement our FCoT-VL-2B and FCoT-VL-8B with CNN for compression. We craft diverse text-oriented understanding tasks, covering OCR-like tasks (i.e.,text recognition, image2markdown, chart2dict\cite{wei2023vary}). We sample a amount of 2 million image-text pairs and obtain fast and stable optimization as shown in Figure~\ref{fig:pretrain_loss}. Compared with scratch training like TextHawk2, which needs 100M data, our FCoT-VL-2B could finish pre-training about 24 hours with 64 GPUs(NPUs) resources.
\begin{table}[ht]
    \setlength{\tabcolsep}{1.5pt}
    \renewcommand{\arraystretch}{1.2}
    \centering
    \resizebox{\linewidth}{!}{
        \begin{tabular}{*{5}c}
            \hline
            \textbf{\makecell{Compress \\ Ratio}}\ & \textbf{DocVQA} & \textbf{ChartQA} & \textbf{InfoVQA} & \textbf{MME} \\
            \hline
            \textbf{0\%}    & 85.90 & 76.24 & 57.66 &   1440         \\
            \textbf{50\%}  &  74.27 & 55.24 & 47.86 & 1355       \\
            \textbf{75\%} & 63.40 & 49.20 & 38.76&1215\\
            \hline
        \end{tabular} 
    }
    \caption{The Performance of pretrain models under different compressing ratio.}
    \label{tab:pretrain_performance}
\end{table}

We discuss the effects of different compressing ratio (50\% and 75\%) during re-alignment pre-training. 
As Table~\ref{tab:pretrain_performance}
lists, we compare the results of pretrain models of FCoT-VL-2B on the benchmark
DocVQA, ChartQA, InfoVQA and MME \cite{fu2024mmecomprehensiveevaluationbenchmark}. Although we employ OCR-like task for re-alignment, the vanilla model's abilities retain to a considerable extent. 
To alleviate the performance drops incurred by compressing visual token, we introduce a post-training stage in Section~\ref{sec:pt} to mitigate performance lost.


\paragraph{Visual token compression modules}
We test the different
compression modules in our FCoT-VL including: (1). Qformer: utilizing one cross-attention to sample fixed visual tokens from ViT backbone. 
(2). CNNs: applying a 1-d convolutional layer with a stride of $2$ for merging tokens.
(3). Pooling: passing
visual tokens via mean pooling operation for 2$\times$ downsampling. To compare the effects of above three methods,
we perform a small-scale SFT on the same re-aligned model with 60k data for convenience of QA evaluation. As Table~\ref{tab:different_adapter} depicts, we find that Qformer suffers from serious performance drops under our data-constrained distillation training, sharing similar conclusion as in previous works\cite{yu2024texthawk2}. In contrast, both CNN and pooling-based architectures exhibit minimal performance degradation compared to the baseline InternVL-2B model. 
Furthermore, FCoT-VL with CNN architecture enjoys rapid loss decline at the beginning of training phases (consuming about 0.1M image-text samples), as illustrated in Figure \ref{fig:pretrain_loss}. Based on these empirical results, we select CNN as the compression module, leading to $2^n$ visual token token downsampling.
\begin{table}[ht]
    \setlength{\tabcolsep}{5pt}
    \renewcommand{\arraystretch}{1.2}
    \centering
    \begin{tabular}{lccc}
        \hline
    \textbf{\makecell{Compress \\ Module}} & \textbf{DocVQA} & \textbf{ChartQA} & \textbf{InfoVQA}\\
        \hline
        \textbf{original}  & 85.90 & 76.24  & 57.66           \\
        \textbf{Qformer}  &  48.23 & 42.32 & 26.36     \\
        \textbf{CNN} & 82.60 & 75.04 & 50.89           \\
        \textbf{Pooling} &  82.44 & 75.43  & 49.83         \\
        \hline
    \end{tabular}  
    \caption{Performance on different visual token compression module at the $50\%$ compressing ratio. }
    \label{tab:different_adapter}
\end{table}

\begin{figure}[ht]
  \includegraphics[width=\linewidth]{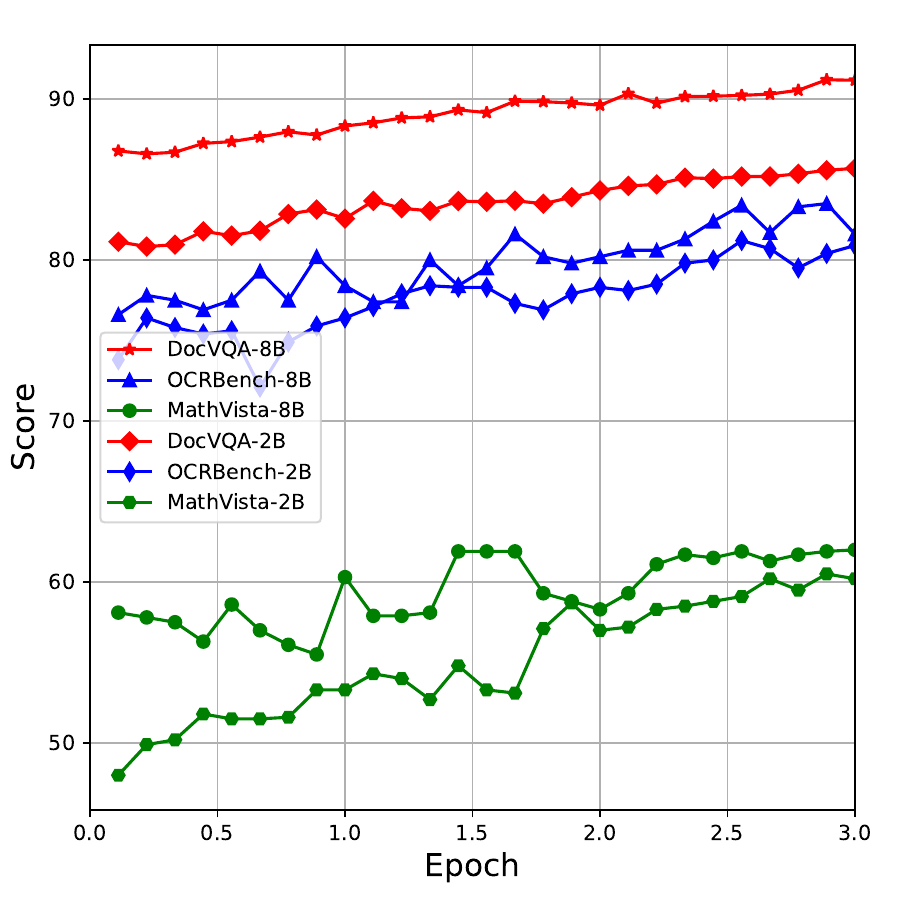}
  \caption {Model performance changes across intermediate training iterations.}
  \label{fig:performancechange}
\end{figure}

\paragraph{Model merge.}
As shown in Figure \ref{fig:performancechange}, we 
observe a performance "see-saw" effect in several benchmarks during different training iterations, motivating us to explore model merging to mitigate this issue.
We compare No Merge (choosing final checkpoint) to three merge strategies with  five intermediate checkpoints, Simple Averaging with equal weight of 0.2, Task Arithmetic \cite{ilharco2022editing} (scaling factor of 0.5) and Sharply-based weighted fusion(Ours). 
As listed in Table~\ref{tab:different_merge_schemes}, Simple Averaging enhances performance compared to no merging across three benchmarks, while Task Arithmetic underperforms in InfoVQA, indicating that Task Arithmetic may not be well-suited for our unified SFT training pipeline. Our method achieves the best results, demonstrating the effectiveness of Sharply-based weight allocation in optimizing checkpoint contributions. Our empirical results suggest that model merging could help tiny-scale VLLMs to achieve heavy post-training.

\begin{figure}[t]
  \includegraphics[width=\linewidth]{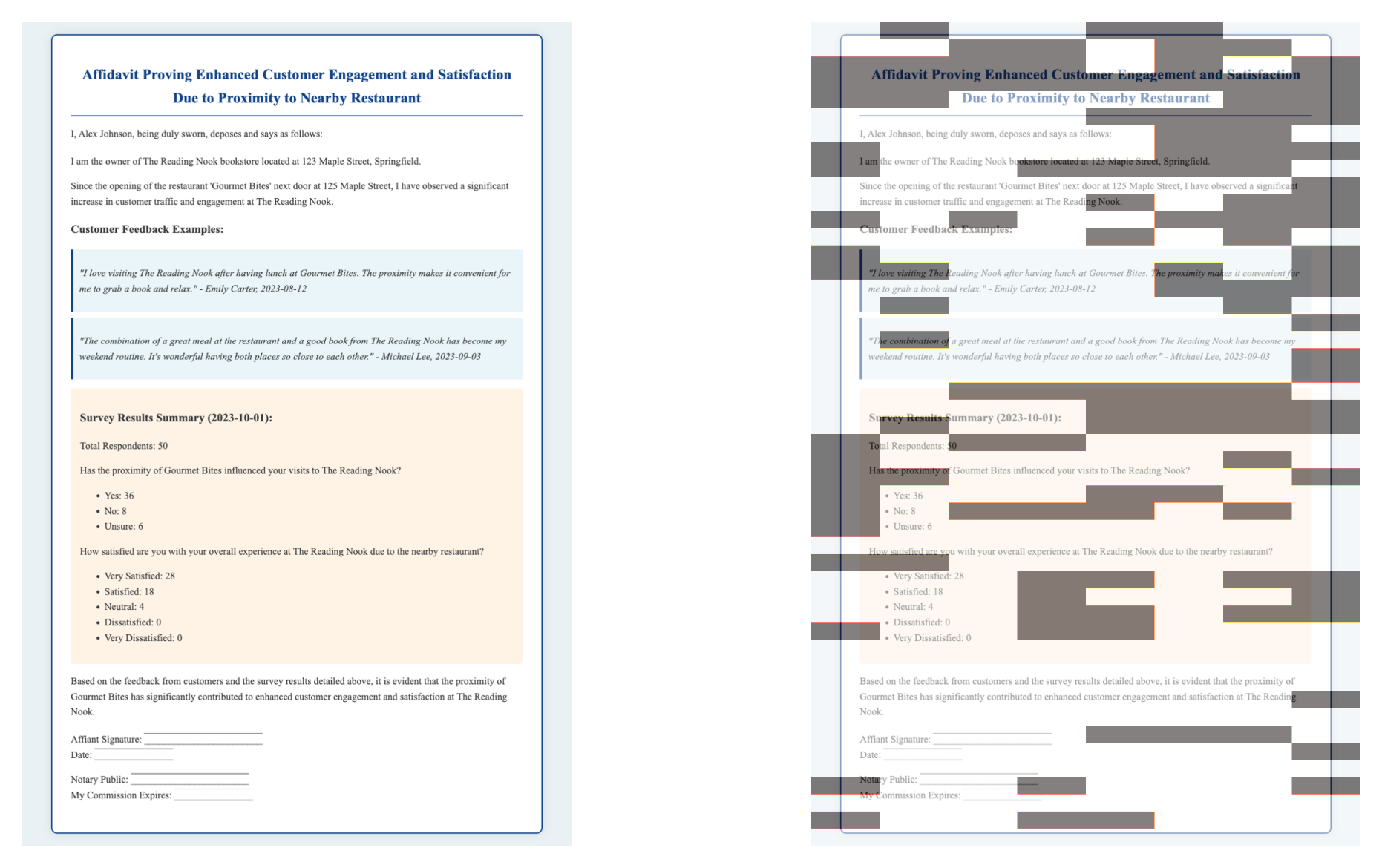} 
  \includegraphics[width = \linewidth]{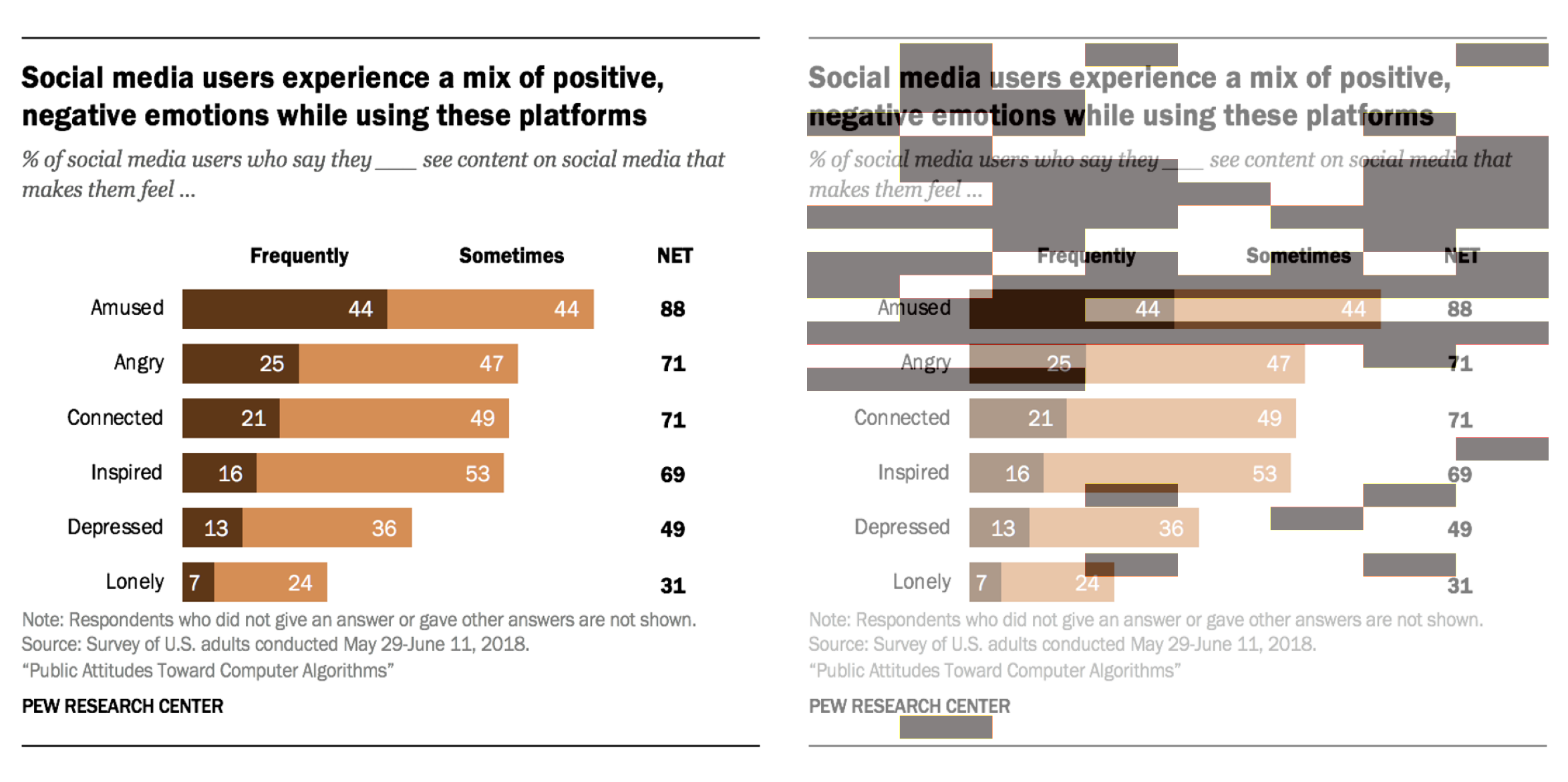} 
  \includegraphics[width=\linewidth]{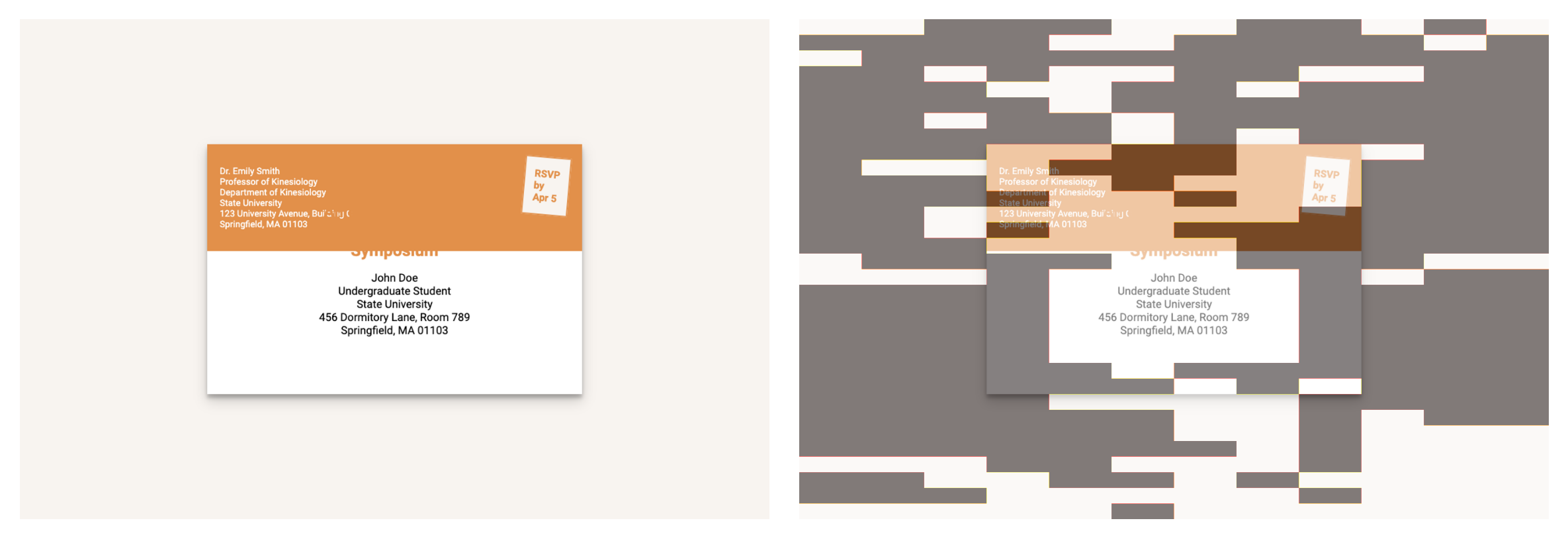} 
  
  \caption { The attention score is calculated from the first layer of the LLM decoder. We use the same prompt: \textbf{please identify the text in the picture} to ask  FCoT-VL-2B(with 50\% reducing).}
  \label{fig:visaulanalysis}
\end{figure}

\begin{table}[ht]
    \setlength{\tabcolsep}{5pt}
    \renewcommand{\arraystretch}{1.2}
    \centering
    \resizebox{\columnwidth}{!}{
        \begin{tabular}{lccccc}
            \hline
            \textbf{Merge Scheme} & \textbf{DocVQA} & \textbf{ChartQA} & \textbf{InfoVQA} \\
            \hline
            \textbf{No Merge}      &       85.12    &      77.31      &       54.43      \\
            \textbf{Simple Averaging} &     86.03      &    78.21        &     55.66        \\
            \textbf{Task Arithmetic}  &      85.31    &     77.43       &      53.14       \\
            \textbf{Ours}             &       \textbf{86.21}  &       \textbf{78.46}     &       \textbf{56.01}      \\
            \hline
        \end{tabular}
    }
    \caption{Performance of Different Merge Schemes on FCoT-VL-2B.}
\label{tab:different_merge_schemes}
\end{table}
\paragraph{Visualization Analysis}
We selected three typical types of text-oriented images, tables, web pages and PPT to visualize the visual token distribution.
As shown in Figure~\ref{fig:visaulanalysis}, the gray part indicates that the model has a lower attention on the concerned pixels. We also find gray part has the overlap with non-text regions, providing an evidence that visual tokens seems redundant even in the 50$\%$ visual token retained.

\paragraph{Inference Speed}\label{timecost}
As shown in Table~\ref{tab:time_cost}, we conduct experiments across three datasets with a single Ascend 910B4 NPU to test the the inference speed of our FCoT-VL models. If we take 75\% ratio as an example, FCoT-VL-2B achieves average 1.5$\times$ faster than baselines at the cost of 5$\%$ performance drops(Table~\ref{tab:main_result}). Furthermore, we observe that the inference efficiency becomes more significant as the LLM backbone is scaled up. Experimental results show that our model offers a cost-effective deployment with a considerably strong performance.
\begin{table}[h]
    \setlength{\tabcolsep}{0.5pt}
    \renewcommand{\arraystretch}{1.2}
    \centering
    \begin{tabular}{cccccccc}
        \hline
        \multirow{2}*{\textbf{\makecell{Model \\ Size}}}& \multirow{2}*{\textbf{\makecell{ Compress\\ Ratio}}} & \multicolumn{2}{c}{\textbf{DocVQA}} & \multicolumn{2}{c}{\textbf{ChartQA}} & \multicolumn{2}{c}{\textbf{InfoVQA}}\\
        ~ & ~ & time & $\delta$ & time & $\delta$ & time & $\delta$ \\
        \hline
        \multirow{3}*{2B} & 0\%  & 782  & -  & 544 & -  & 868 & - \\
        ~ & 50\% & 598 & 1.3$\times$ & 467 & 1.2$\times$ & 614 & 1.4$\times$ \\
        ~ & 75\% & 553 & \textbf{1.4$\times$} & 346 & \textbf{1.6$\times$} & 600 & \textbf{1.4$\times$}\\
        \hline
        \multirow{3}*{8B} & 0\% &1279 & -  & 704 &-  &1457 & - \\
        ~ & 50\% &838 & 1.5$\times$ & 544 &1.3$\times$ & 857 & 1.7$\times$\\
        ~ & 75\% &673 & \textbf{1.9$\times$} & 474 &\textbf{1.5$\times$} & 614 & \textbf{2.4$\times$}\\
        \hline
    \end{tabular}
    \caption{Inference time experiments on a single Ascend 910B4 NPU. Time is measured in milliseconds, and $\delta$ denotes the reduction ratio.}
    \label{tab:time_cost}
\end{table}

\section{Conclusion}
In this paper, we introduce FCoT-VL, a novel method designed to efficiently compress Vision-Language Large Models (VLLMs) by reducing redundant visual tokens with minimal computational resources, while maintaining or even enhancing model performance. FCoT-VL significantly reduces the number of visual tokens, achieving notable performance improvements. Furthermore, FCoT-VL is highly resource-efficient, requiring minimal data and NPU resources. The method demonstrates strong compatibility with various compression modules, all of which perform well in conjunction with FCoT-VL. Extensive experiments across multiple benchmarks confirm that FCoT-VL excels in tasks requiring fewer visual tokens, including text-oriented tasks, even when computational resources are limited.

\section{Limitations}
(1).We only focus on text-oriented tasks that require high-resolution settings and obtain lossless compression with the ratio of 50\%. However, due to resource constraints, our approach does not extend to other image modalities, such as natural scenes or medical imaging. (2). Although our fixed compression ratios (i.e., 50\% and 75\%) are efficiently implemented, this setting performs well in most cases. However, it shows a slight performance drop when applied to extremely high-resolution tasks, such as infoVQA.

\bibliography{custom}

\appendix

\section{Appendix}
\label{sec:appendix}

\subsection{Training settings}
Our FCoT-VL was trained in two distinct stages: re-alignment and post-trian. 
As shown in Table~\ref{tab:training_settings}, we present the training details of FCoT-VL in different stages. The details are as follows:

For both stages, we train models with 64 ascend 910 NPUs with the packed batch size is set to 512.
In the re-alignment pre-training, we employ a 2 million image-text pairs to learn the projector and compress module. This allows the VLLMs to re-align the compressed visual token with the language token space. Specifically, we craft the optimization tasks of recognizing text in document images and converting charts and tables into pythondict/markdown format. We set the training epoch as 1, which requires approximately 48 hours using 64 NPUs for 2B scale. In the subsequent instruction tuning phase, we make all parameters of FCoT-VL learnable and keep most of the settings unchanged, except context length, training data and training epochs. 


  \begin{table}[htbp]
    \renewcommand{\arraystretch}{1.2}
    \centering
    \resizebox{\columnwidth}{!}{
        \begin{tabular}{l|cc}
            \hline
            \textbf{Settings} & \textbf{Re-alignment} & \textbf{Post-train} \\
            \hline
            \rowcolor{gray!15}
            Trainable & \makecell{Projector,\\ Compress Module} & Full Parameters \\
            Packed Batch Size & 512 & 512  \\
            \rowcolor{gray!15}
            Learning Rate & $1e^{-5}$ & $1e^{-5}$ \\
            Context Length & 4096 & 5120 \\
            \rowcolor{gray!15}
            Image Tile Threshold & 12 & 12 \\
            ViT Drop Path & 0.1 & 0.1 \\
            \rowcolor{gray!15}
            Weight Decay & 0.01 & 0.01  \\
            Training Epochs & 1 & 3 \\
            \hline
            \rowcolor{gray!15}
            Dataset & Pre-train & Fine-tune \\
            Training Examples  & $\sim2M$ & $\sim4.5M$ \\
            \hline
        \end{tabular}
    }
    \caption{Detailed Training settings for InternVL2-2B and InternVL2-8B.}
    \label{tab:training_settings}
\end{table}


\subsection{Model Capabilities and Qualitative Examples}
In this section, we present some practical examples of our FCoT-VL.
\begin{figure*}[htbp]
    \centering
    \includegraphics[width=\textwidth]{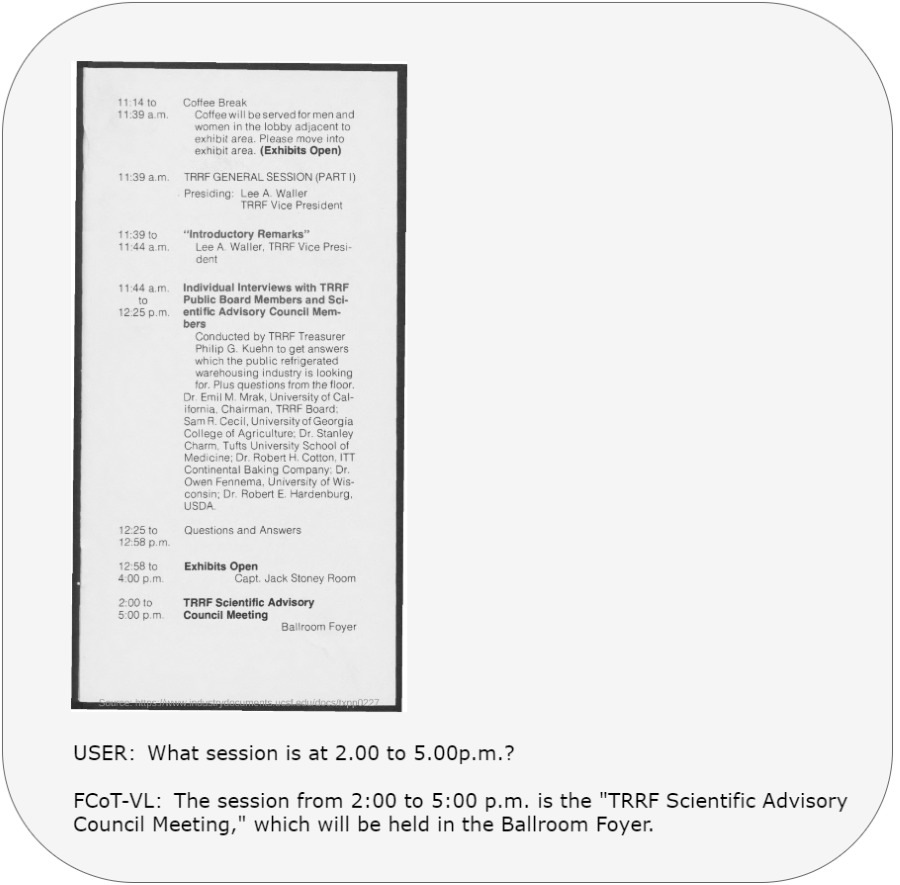}  
    \caption{The model excels in understanding scheduling-related queries. Image source:\cite{mathew2021docvqa}}  
    \label{ex1}  
\end{figure*}

\begin{figure*}[htbp]
    \centering
    \includegraphics[width=\textwidth]{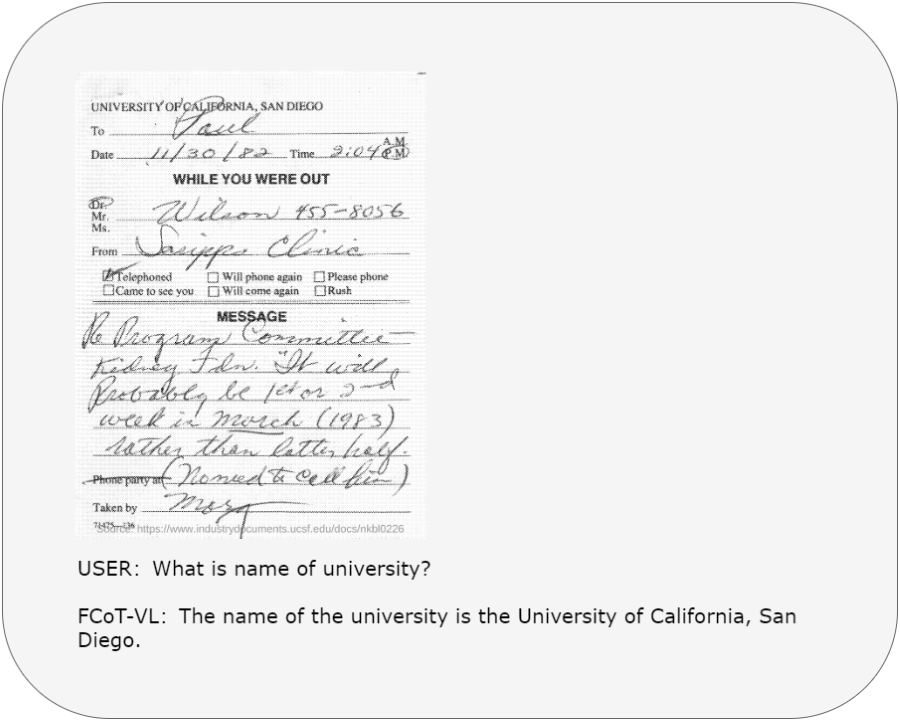}  
    \caption{The model demonstrates excellence in recognizing handwritten text in emails. Image source:\cite{mathew2021docvqa}}  
    \label{ex2}  
\end{figure*}

\begin{figure*}[htbp]
    \centering
    \includegraphics[width=\textwidth]{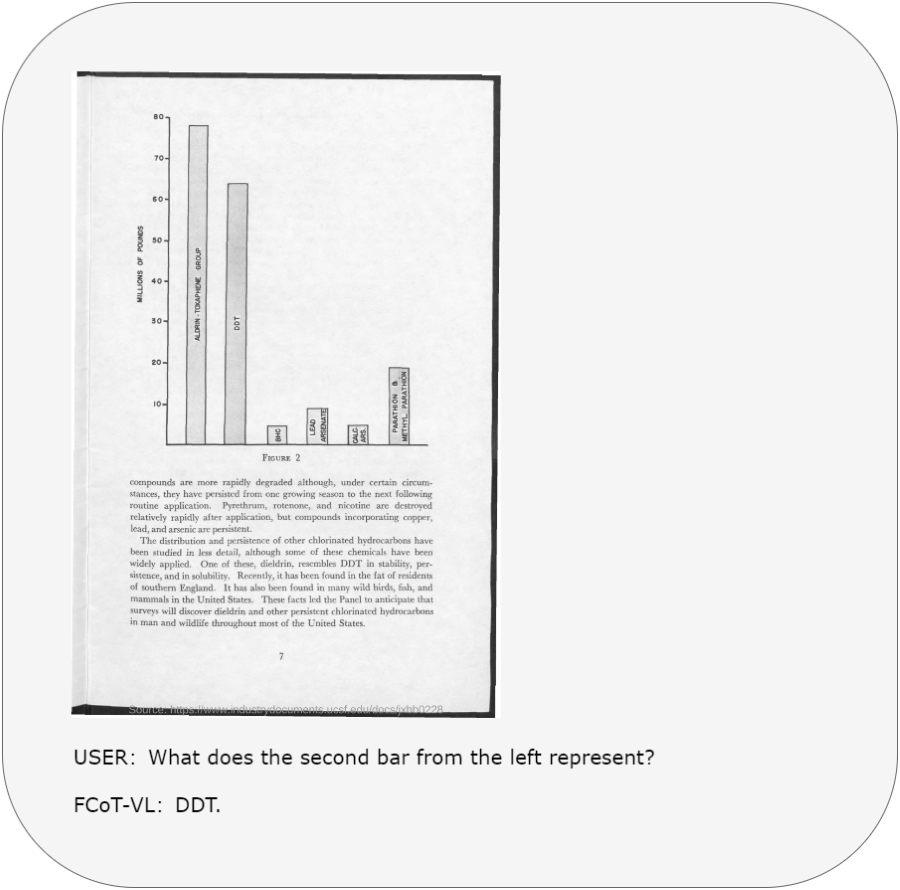}  
    \caption{The model demonstrates excellence in recognizing printed text and images in books. Image source:\cite{mathew2021docvqa}}  
    \label{ex3}  
\end{figure*}

\begin{figure*}[htbp]
    \centering
    \includegraphics[width=\textwidth]{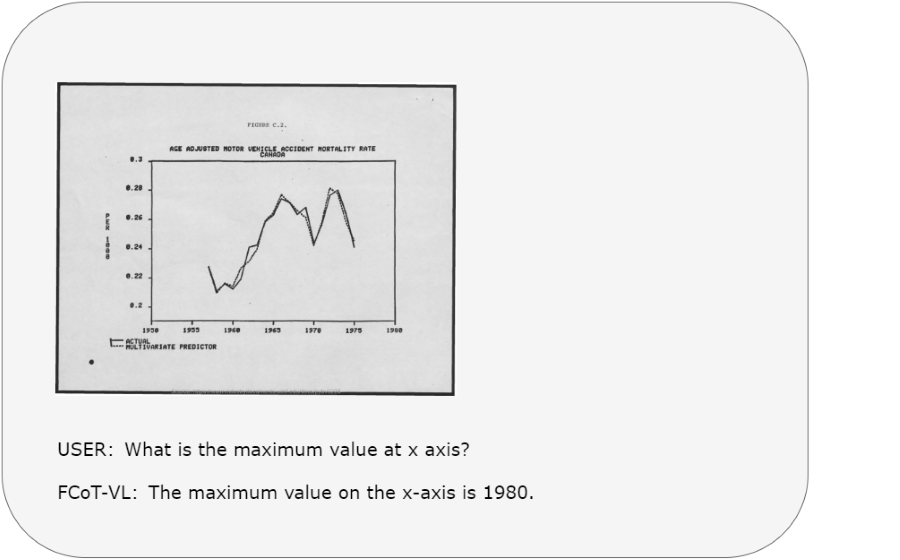}  
    \caption{The model displays an adeptness in understanding line charts. Image source:\cite{mathew2021docvqa}}  
    \label{ex4}  
\end{figure*}

\begin{figure*}[htbp]
    \centering
    \includegraphics[width=\textwidth]{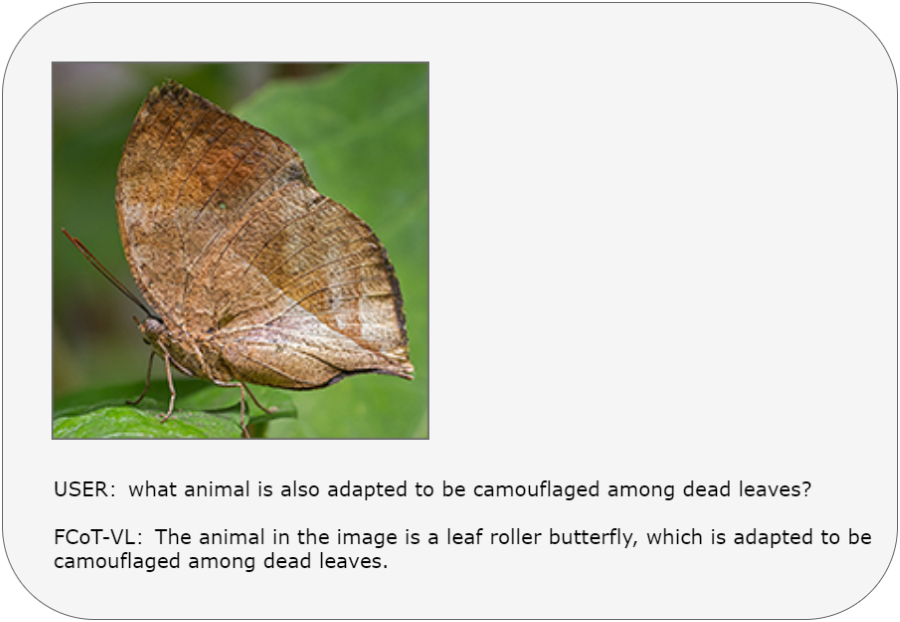}  
    \caption{The model displays an adeptness in understanding images of natural animals. Image source:\cite{lu2022scienceqa}}  
    \label{ex5}  
\end{figure*}

\begin{figure*}[htbp]
    \centering
    \includegraphics[width=\textwidth]{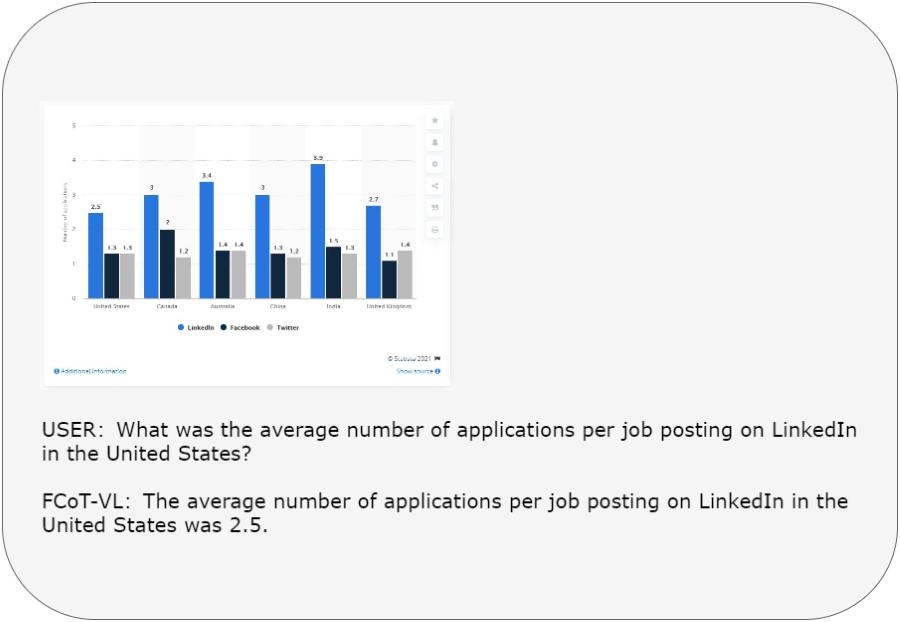}  
    \caption{The model displays an adeptness in understanding bar charts. Image source:\cite{masry2022chartqa}}  
    \label{ex8}  
\end{figure*}

\begin{figure*}[htbp]
    \centering
    \includegraphics[width=\textwidth]{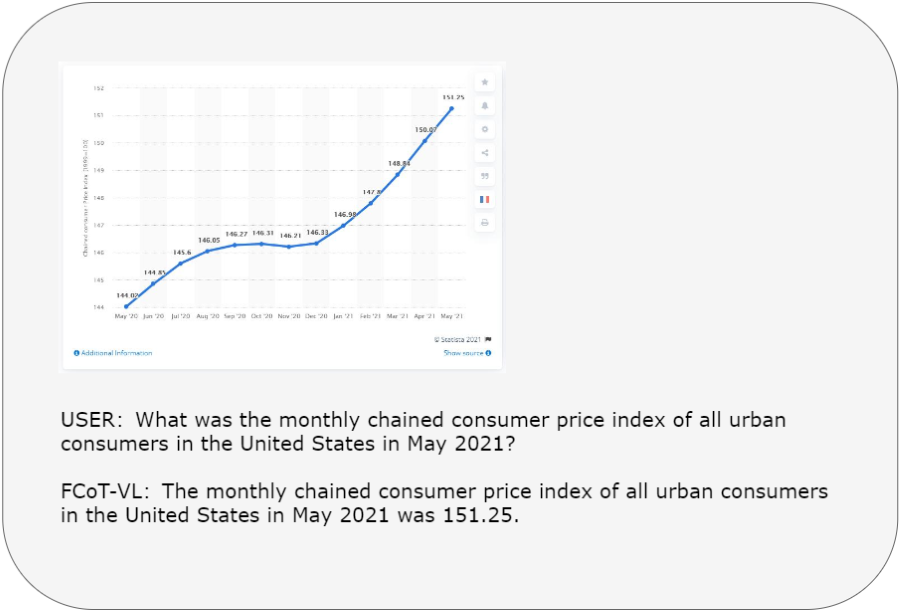}  
    \caption{The model displays an adeptness in understanding curve charts. Image source:\cite{masry2022chartqa}}  
    \label{ex9}  
\end{figure*}

\begin{figure*}[htbp]
    \centering
    \includegraphics[width=\textwidth]{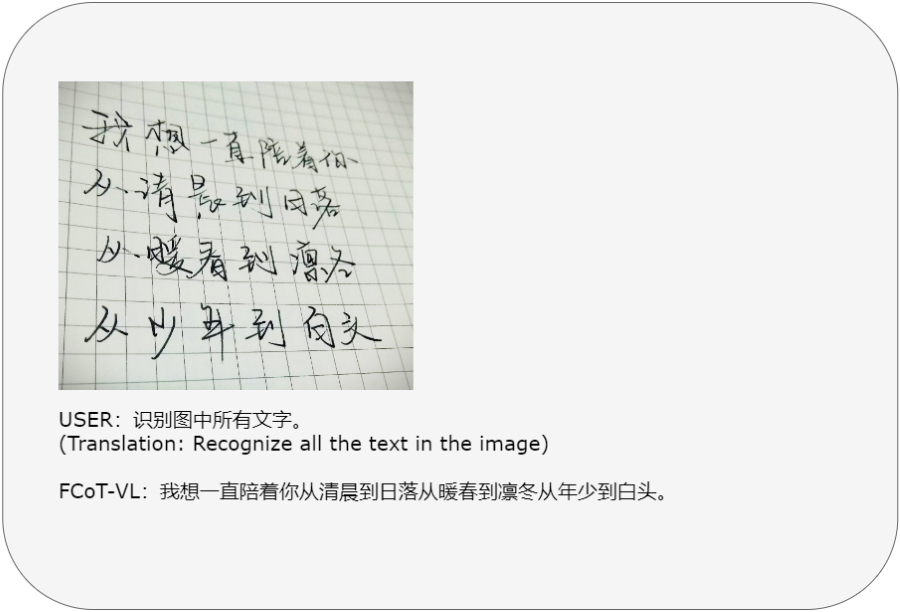}  
    \caption{The model displays an adeptness in recognizing handwritten Chinese characters.}  
    \label{ex10}  
\end{figure*}

\begin{figure*}[htbp]
    \centering
    \includegraphics[width=\textwidth]{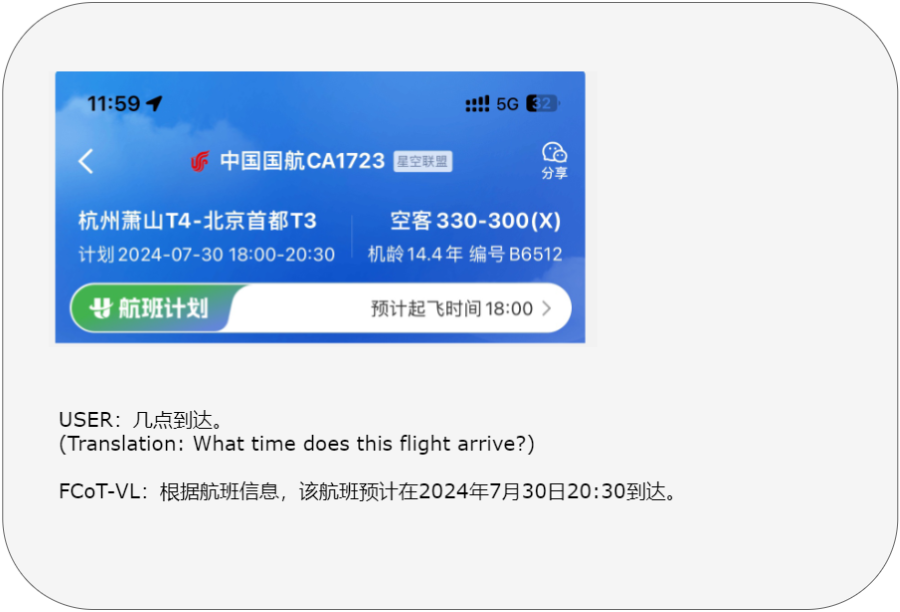}  
    \caption{The model displays an adeptness in understanding Chinese flight ticket information.Image source:\cite{wang2024qwen2vl}}  
    \label{ex11}  
\end{figure*}

\begin{figure*}[htbp]
    \centering
    \includegraphics[width=\textwidth]{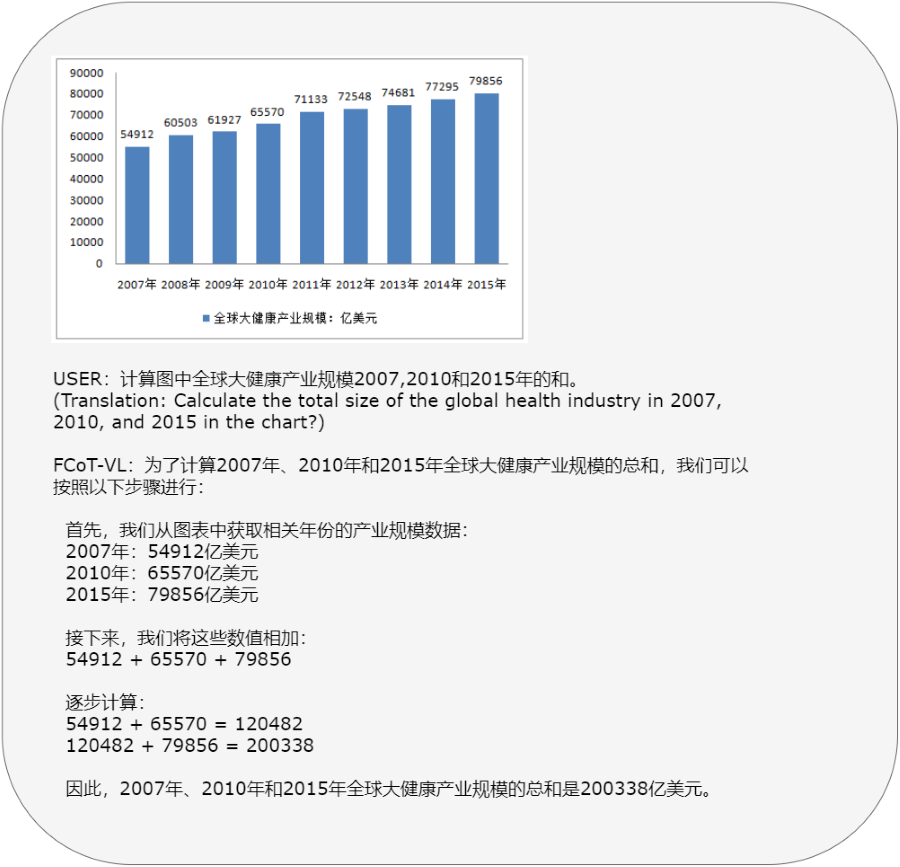}  
    \caption{The model displays an adeptness in calculating information from Chinese bar charts. }  
    \label{ex12}  
\end{figure*}

\begin{figure*}[htbp]
    \centering
    \includegraphics[width=0.25\textwidth]{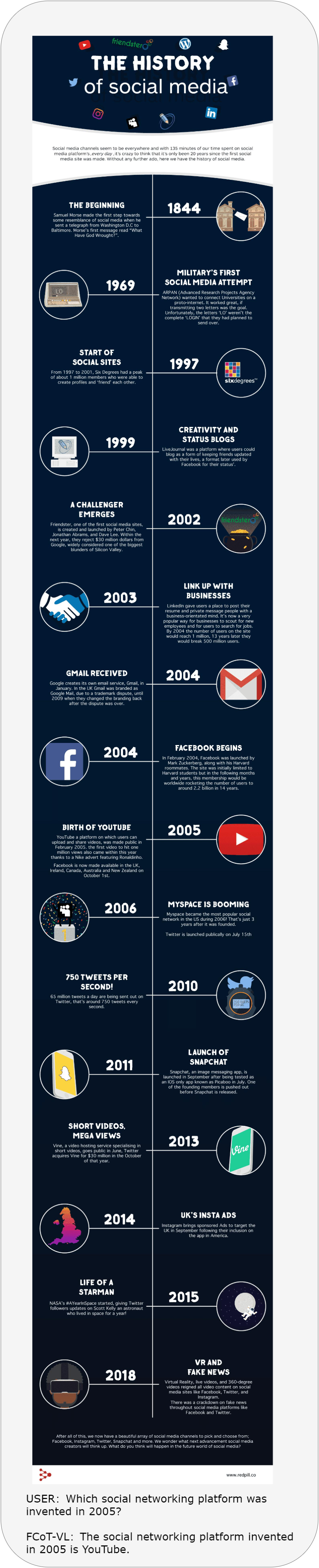}  
    \caption{The model displays an adeptness in understanding posters with dense information. Image source:\cite{mathew2022infographicvqa}}  
    \label{ex6}  
\end{figure*}

\begin{figure*}[htbp]
    \centering
    \includegraphics[width=0.23\textwidth]{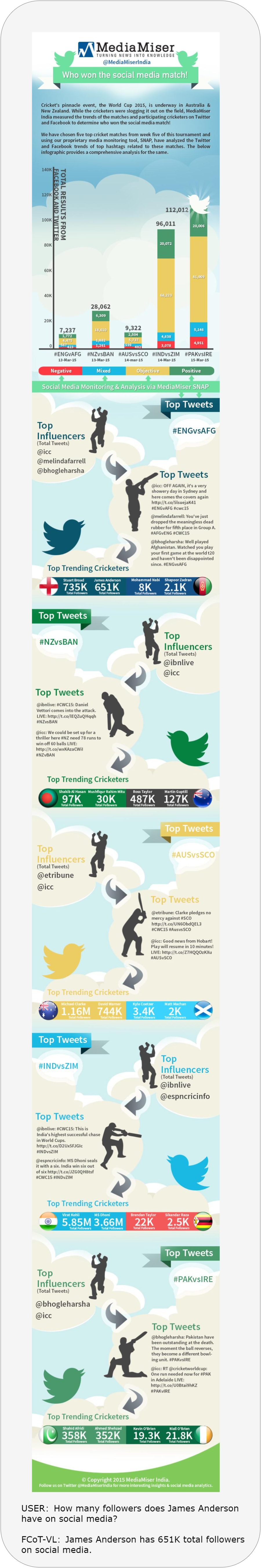}  
    \caption{The model displays an adeptness in understanding posters with intertwined text and images. Image source:\cite{mathew2022infographicvqa}}  
    \label{ex7}  
\end{figure*}

\end{document}